\documentclass{tlp}

\usepackage{latexsym}
\usepackage{amssymb}
\usepackage{amsmath}
\usepackage{booktabs}
\usepackage{enumitem}
\usepackage{graphicx}
\usepackage{color}
\usepackage{hyperref}

\newtheorem{theorem}{Theorem}

\newtheorem{proposition}[theorem]{Proposition}

\newtheorem{example}{Example}

\newcommand{\BibTeX}{B\kern-.05em{\sc i\kern-.025em b}\kern-.08em\TeX}
\usepackage{mathtools}
\usepackage{xspace}
\usepackage{multirow}
\usepackage{graphics}
\usepackage[export]{adjustbox}

\usepackage{lscape}
\usepackage{url}
\def\sm{{SM}\xspace}

\def\sr{{SR}\xspace}
\def\sri{{SRI}\xspace}
\def\srt{{SRT}\xspace}
\def\srti{{SRTI}\xspace}

\def\srtiasp{{\sc SRTI-ASP}\xspace}

\def\k-cp{{$k$-connected pairs}\xspace}

\def\lar{\leftarrow}
\def\ba{\begin{array}}
\def\ea{\end{array}}
\def\beq{\begin{equation}}
\def\eeq#1{\label{#1}\end{equation}}
\def\no{\ii{not}}
\def\ii#1{\hbox{\it #1\/}}

\def\ie{i.e., }
\def\eg{e.g., }

\usepackage{todonotes}

\begin{document}

\lefttitle{Fidan and Erdem}

\jnlPage{1}{8}
\jnlDoiYr{2021}
\doival{10.1017/xxxxx}

\title[Finding Personalized Good-Enough Solutions to Unsatisfiable \sr]{Finding Personalized Good-Enough Solutions to \\ Unsatisfiable Stable Roommates Problems}

\begin{authgrp}
\author{\gn{M\"UGE FIDAN} \and \gn{ESRA ERDEM}}
\affiliation{Faculty of Engineering and Natural Sciences, Sabanci University, Istanbul, Turkey \\ \emails{mugefidan@sabanciuniv.edu}{esraerdem@sabanciuniv.edu}}
\end{authgrp}

\history{\sub{xx xx xxxx;} \rev{xx xx xxxx;} \acc{xx xx xxxx}}

\maketitle

\begin{abstract}
The Stable Roommates problems are characterized by the preferences of agents over other agents as roommates. A solution is a partition of the agents into pairs that are acceptable to each other (\ie they are in the preference lists of each other), and the matching is stable (\ie there do not exist any two agents who prefer each other to their roommates, and thus block the matching). Motivated by real-world applications, and considering that stable roommates problems do not always have solutions, we continue our studies to compute ``good-enough" matchings. In addition to the agents' habits and habitual preferences, we consider their networks of preferred friends, and introduce a method to generate personalized solutions to stable roommates problems. We illustrate the usefulness of our method with examples and empirical evaluations.
\end{abstract}

\begin{keywords}
    stable roommates problem, answer set programming, declarative problem solving
\end{keywords}

\section{Introduction}
The Stable Roommates problem~\citep{galeShapley1962} (\sr) is a matching problem (well-studied in Economics and Game Theory) characterized by the preferences of an even number~$n$ of agents over other agents as roommates: each agent ranks all others in strict order of preference. A solution to \sr is then a partition of the agents into pairs that are {\em acceptable} to each other (i.e., they are in the preference lists of each other), and the matching is {\em stable} (i.e., there exist no two agents who prefer each other to their roommates, and thus {\em block} the matching).

Motivated by real-world applications, variations of \sr have been investigated. For instance, students may find it difficult to rank a large number of alternatives in strict order of preference. With such motivations, \sr has been studied with incomplete preference lists (\sri)~\citep{gusfield89}, with preference lists including ties (\srt)~\citep{RONN1990}, and with incomplete preference lists including ties (\srti)~\citep{irving2002}. In this study, we focus on \srti.

As first noted by 
\cite{galeShapley1962}, unlike the well-known Stable Marriage problem (\sm), there is no guarantee to find a solution to every \sr problem instance (i.e., there might be no stable matching).  In real-world applications, the preference lists are often too short or empty (e.g., for freshmen students). Thus, for \sri and \srti instances, it gets harder to find a stable matching: students who are not in the preference lists of each other are considered unacceptable to each other, and thus they cannot be matched. In such cases, we still need to find a matching that is ``good-enough".

To find good-enough matchings, existing approaches consider more general variations of \sr. For instance, Almost \sr~\citep{abraham2005} aims to find a matching that minimize the total number of blocking pairs (i.e., pairs of agents who prefer each other to their roommates).
\cite{Gardenfors1975} consider ``popularity" instead of stability and aims to find a matching that maximizes the number of agents who are better-off in this matching compared to other matchings.
\cite{Thakur2021} introduces the property of ``majority stability" to find a matching preferred by more than half of all agents over other matchings.
\cite{aziz2023} generalize the idea of majority stability by proposing ``$k$-stability" which aims to find a matching preferred by at least~$k$ out of the~$n$ agents.
To relax restriction of popularity, \cite{Kavitha2023} consider the notion of ``semi-popularity" to find a matching which is undefeated by a majority of matchings.
\cite{Afacan2016} introduce the notion of ``sticky stability" to accommodate appeal costs in school placements, and allows priority violations when the rank difference between the claimed and the received objects are less than a certain threshold. In all these studies, every stable matching (if one exists) is also one of the targeted matchings, but not vice versa.

We propose a different (but orthogonal) approach to find good-enough solutions, by extending the preference lists of agents with ``suitable" candidates identified by further available knowledge.  For instance, in our earlier work~\citep{fidan2021}, we have introduced knowledge-based methods for \srti, that identify suitable candidates by considering the habits (e.g., smoking, studying) of students, the preferences over the habits of their roommates, and the preferences and the constraints of the colleges (e.g., to increase diversity and inclusion). We have observed the usefulness of this approach in computing not only good-enough matchings that are stable with respect to extended lists, but also more personalized, more diverse and/or more inclusive matchings.

In this study, motivated by the promising results of our earlier studies and further observations we have made in real-world applications with the help of different parties involved, we propose a new method to further extend preference lists to be able to find a matching when there is no stable matching. In particular, based on the observations, students often make new friends by meeting friends of their friends, our method extends the preference lists further by considering the networks of agent's preferred friends.
At this point, the curious reader may think rightfully that not every student would agree with the preference lists of their friends, but actually many of the students do prefer as roommates a preferred friend's friend if they cannot be assigned to any of their preferred friends. Indeed, according to a poll survey, as shown in Fig.~\ref{poll-survey},
conducted with 101 students at Sabanci University, the majority ($75.2\%$) of the participants have expressed their preference of a roommate from the preference lists of their friends, if they cannot be assigned to any of their preferred friends. Hence, based on these results, extending the preference lists, based on their networks of preferred friends, is beneficial to align with student's expectations for assigning a roommate. Based on this observation, our method further extends the preference list of an agent~$x$ by including every agent~$y$ who is not in~$x$'s preference list but ``$k$-connected" to~$x$ (i.e.,~$x$ and~$y$ are connected by a chain of~$k$ PFOAF--preferred friend of a friend--relations).
\begin{figure}[h!]
\center
\noindent\fbox{\begin{minipage}{0.8\textwidth}
{\it Imagine that you're about to start a new semester at the university, and you apply for a double room in a dormitory. You know that some of your friends will apply for a double room as well, so you also specify a list of your friends that you would like to roommate with. After the dormitory system collects all applications, it will try to find the ``best'' roommate for every applicant considering their preference lists, and place the applicants in double rooms. \\

However, finding a ``best'' roommate for each applicant is a challenging task, and there is a chance that the dormitory system cannot match you with any of your friends on your preference list. In that case, would you prefer to roommate with someone from the preference lists of your friends?

\smallskip

(a) Yes, I prefer this.

\smallskip

(b) It does not matter, anyone will do.}
\end{minipage}}
\label{poll-survey}
	 	\caption{A poll survey given to a set of students at Sabanci University, to find out whether the student prefers as roommate (a) someone from the preference lists of their preferred friends, or (b) anyone. }
\end{figure}


Also, based on our interviews (cf. the two interviews presented in Appendix A)
we consider that some students may request not to be matched with some other students (e.g., previous roommates who they could not get along). Our method utilizes this knowledge about ``forbidden pairs" as well, while identifying suitable candidates to extend the preference lists.


We illustrate the benefits of utilizing different types of knowledge to extend the preference lists, with examples. Also, we implement our method by utilizing the expressive languages and efficient solvers of Answer Set Programming (ASP)~\citep{Niemelae99,MarekT99,Lifschitz02,BrewkaEL16}, and evaluate its usefulness by a comprehensive set of experiments.

\section{Preliminaries}

In this study, we consider the 
definitions of~\srti~\citep{erdem2020} and its knowledge-based extentions (\ie Personalized-\srti)~\citep{fidan2021}, as well as the ASP formulations to solve these problems, as in our earlier studies. 

\subsection{\srti: Stable Roommates problem with Ties and Incomplete Lists}
{\label{sec:srti}}

Let~$A$ be a finite set of agents. For every agent~$x\in A$, let
$A_x \subseteq A \backslash \{x\}$ be a set of agents that are {\em acceptable} to~$x$ as roommates. For every~$y$ in~$A_x$, we assume that~$x$ prefers~$y$ as a roommate compared to being single.

Let $\prec_{x}$ be a partial ordering of~$x$'s preferences over~$A_x$ where incomparability is transitive. We refer to~$\prec_{x}$ as agent~$x$'s preference list. For two agents~$y$ and~$z$ in~$A_x$, we denote by~$y \prec_{x} z$ that~$x$ prefers~$y$ to~$z$. In this context, ties correspond to indifference in the preference lists: an agent~$x$ is {\em indifferent} between the agents~$y$ and~$z$, denoted by~$y \sim_{x} z$, if~$y \not \prec_{x} z$ and~$z \not \prec_{x} y$. We denote by~$\prec$ the collection of all preference lists.

A \emph{matching} for a given \srti instance is a function
${M: A \mapsto A}$ such that, for all~$\{x,y\} \subseteq A\times A$ such that~$x\in A_y$ and~$y\in A_x$,~$M(x)=y$ if and only if~$M(y)=x$. If agent~$x$ is mapped to itself, we then say he/she is \emph{single}.

A matching~$M$ is {\em blocked} by a pair~$\{x, y\} \subseteq A \times A$ ($x\neq y$) if the following hold:
both agents~$x$ and~$y$ are acceptable to each other,
$x$ is single with respect to~$M$, or~$y \prec_{x} M(x)$, and
$y$ is single with respect to~$M$, or~$x \prec_{y} M(y)$.

A matching for \srti is called {\em stable} if it is not blocked by any pair of agents. 

We can declaratively solve \srti using ASP as described in our earlier work~\citep{erdem2020}.
The input~$I=(A,\prec)$ of an \srti instance is formalized by a set~$F_I$ of facts using atoms of the forms~$\ii{agent}(x)$ (``$x$ is an agent in~$A$'') and~$\ii{prefer2}(x,y,z)$ (``agent~$x$ prefers agent~$y$ to agent~$z$, i.e.,~$y \prec_{x} z$''). For every agent~$x$,  for every~$y\in A_x$, we also add facts of the form~$\ii{prefer2}(x,y,x)$ to express that~$x$ prefers~$y$ as a roommate instead of being single. We recursively define the transitive closure of this preference relation:
$$
\ba l
\ii{prefer}(x,y,z) \lar \ii{prefer2}(x,y,z). \\
\ii{prefer}(x,y,z) \lar \ii{prefer2}(x,y,w), \ii{prefer}(x,w,z).
\ea
$$
and acceptability of agents:
\beq
\ba l
\ii{accept}(x,y) \lar \ii{prefer}(x,y,\_). \\
\ii{accept}(x,y) \lar \ii{prefer}(x,\_,y). \\
\ii{accept2}(x,y) \lar \ii{accept}(x,y), \ii{accept}(y,x).
\ea
\eeq{eq:P-accept}

The output~$M: A \mapsto A$ of an \srti instance is characterized by atoms of the form~$\ii{room}(x,y)$ (``agents~$x$ and~$y$ are roommates'').  For every agent~$x$, exactly one mutual acceptable agent~$y$ is nondeterministically chosen as~$M(x)$:
$$
\ba l
1\{\ii{room}(x,y){:} \ii{agent}(y), \ii{accept2}(x,y)\}1 \lar \ii{agent}(x). \\
\lar \ii{room}(x,y), \no\ \ii{room}(y,x).
\ea
$$
and the stability of the generated matching is ensured:
$$
\lar \ii{block}(x,y) \qquad (x\neq y) .
$$
\noindent where atoms of the form~$\ii{block}(x,y)$ describe the blocking pairs.

\subsection{Personalized-\srti: \srti with Personalized Criteria}


Let~$B$ be a finite list~$\langle b_{1},b_{2},\dots,b_{k} \rangle$ of criteria. For every~$b_{i} \in B$, let~$C_i$ be a finite list~$\langle c_{i1},c_{i2},\dots,c_{im} \rangle$ of its choices for~$b_{i}$, ordered with respect to a ``closeness'' measure.

Let~$f$ be a function that maps an agent~$x\in A$ and a criterion~$b_{i} \in B$ to a positive integer~$j$ ($1\leq j \leq |C_{i}|$), describing the choice~$c_{ij}$ of the agent~$x$.
For every agent~$x$, $x$'s {\em (preference) profile}~$P_{x}=\langle f(x,b_{1}),f(x,b_{2}),\dots,f(x,b_{k})\rangle$ characterizes its choices of for each criterion in~$B$, respectively. 

Every criterion in~$B$ may have a different importance for each agent. For that, we introduce a weight function~$w$ that maps an agent~$x\in A$
and a criterion~$b_{i} \in B$ to a non-negative integer such that~$w(x,b_{i})$ denotes the importance of the criterion~$b_{i}$ for~$x\in A$. For every agent~$x\in A$, let us denote by the weight list~$W_{x}=\langle w(x,b_{1}),w(x,b_{2}),\dots,w(x,b_{k})\rangle$ the respective weights of criteria in~$B$ for~$x$. 

For every agent~$x\in A$, with a profile~$P_x$ and a weight list~$W_x$, let us denote the criteria of the same weight~$u>0$ and the agent~$x$'s choices for them, by a nonempty set~$E_u$ of tuples:
$E_u = \{(f(x,\pi_{i}),\pi_i)\ |\ u=w(x,\pi_i)>0,\ \pi_i \in \{b_{1},b_{2},\dots,b_{k}\}\}.$
Then, for every agent~$x\in A$, we define a {\em sorted profile}~$P_{x}^{'}$ for~$x$,
with respect to~$P_{x}$ and~$W_x$:
$P_{x}^{'}=\langle E_{u_1}, E_{u_2}, \dots, E_{u_m} \rangle$
where~$m \leq k$, and, for each~$i$ ($1 {\leq} i {<} m$),~$u_i > u_{i+1}$.

For every agent~$y\in A \backslash A_{x}$ (i.e.,~$y$ is not acceptable to~$x$), $y$ is {\em choice-acceptable} to~$x$ if there exists some criterion~$b_{i}\in B$ where~$w(x,b_i)>0$ such that~$f(y,b_{i})=f(x,b_{i})$.
%
For every agent~$x$ with a sorted profile~$P_{x}^{'}=\langle E_{u_1}, E_{u_2}, \dots, E_{u_m} \rangle$ ($m \leq k$), for every two agents~$y$ and~$z$ that are choice-acceptable to~$x$, the agents~$y$ and~$z$ are {\em choice-equal} for~$x$ relative to the first~$j$ sets~$E_{u_1}, E_{u_2}, \dots, E_{u_j}$ in~$P_{x}^{'}$ (denoted~$y =_{x} z |_j$) if
$j=0$, or
$j>0$,~$y =_{x} z |_{j-1}$, and, for every~$(f(x,\pi_{i}),\pi_i)\in E_{u_j}$,~$f(x,\pi_{i}) = f(y,\pi_{i}) = f(z,\pi_{i})$.

We say that~$x$ {\em prefers}~$y$ to~$z$ {\em with respect to a sorted profile}~$P_{x}^{'}$ (denoted~$y \prec'_{x} z$) if, for some~$j>0$, 
$y =_{x} z |_{j-1}$, and
$|\{\pi_i |\ (f(x,\pi_{i}),\pi_i){\in} E_{u_j},\ f(x,\pi_{i}) {=} f(y,\pi_{i})\}|\ >\ $ 
$|\{\pi_i |\ (f(x,\pi_{i}),\pi_i){\in} E_{u_j},\ f(x,\pi_{i}) {=} f(z,\pi_{i})\}|.$

A {\em criteria-based personalized preference list}~$\prec_{x}^{'}$ is a partial ordering of~$x$'s preferences over~$A'_x$ with respect to a sorted profile~$P_{x}^{'}$, where incomparability is transitive.
%
An {\em extended preference list}~$\prec_{x}^{''}$ is obtained by concatenating~$\prec_{x}$ and $\prec_{x}^{'}$, depending on the importance given to these two types of lists.
{\em Personalized-\srti} is then characterized by~$(A,\prec^{''})$ where~$A$ is a finite set of agent, and~$\prec^{''}$ is the collection of the extended preference list of each agent~$x~\in~A$. 

\begin{table}[t!]
\caption{A Personalized-\srti instance, characterized by a set of students~$x$, and their extended preference lists~$\prec_{x}^{''}$.}
\label{instance1}
{\tablefont
{\begin{tabular}{@{\extracolsep{\fill}}cccc}
\topline
Student~$x$  & $\prec_{x}$  & $\prec_{x}^{'}$  & $\prec_{x}^{''}$ \\
& (given) & (inferred) & (extended)
\midline
$a$   & $\langle e \rangle$   & $\langle b \rangle$  & $\langle e, b \rangle$ \\
$b$   & $\langle e \rangle$   & $\langle \rangle$    & $\langle e \rangle$ \\
$c$   & $\langle b \rangle$   & $\langle \{a,e\} \rangle$  & $\langle b, \{a,e\} \rangle$ \\
$d$   & $\langle b \rangle$   & $\langle \rangle$  & $\langle b \rangle$  \\
$e$   & $\langle d \rangle$   & $\langle \rangle$  & $\langle d \rangle$
\botline
\end{tabular}}}
\end{table}

\begin{example}
Suppose that a set~$\{a,b,c,d,e\}$ of freshman students {applies} for an accommodation at a dormitory by filling out a questionnaire, like the one shown in Table 4 of~\citep{fidan2021}. The roommate questionnaires request from the students to provide their preferences over others, e.g., by considering first impressions if they have a chance to meet. Table~\ref{instance1} shows preferences~$\prec_{x}$ of these students, extracted from these questionnaires.  Here, the \srti instance~$I=(A,\prec)$ does not have a stable matching since there is no pair of students who are acceptable to each other.

So, let us identify some suitable candidates relative to the habits (e.g., smoking, studying) of students, and their priorities and preferences over the habits of their roommates, and then extend the preference lists by including these suitable candidates.  The ordering~$\prec_{x}^{'}$ of the suitable candidates is shown in Table~\ref{instance1}. Considering the aggregate preference list~$\prec_{x}^{''}=\prec_{x} + \prec'_{x}$, the Personalized-\srti instance~$I'=(A,\prec^{''})$ does not have a stable matching.

\end{example}

\section{$k$-Personalized-\srti: Personalized-\srti with~$k$-connection}

Students often meet other students through mutual friends, e.g., the friends of their friends,
and sometimes get along with each other. Based on this observation,
we propose a new method to further extend preference lists to be able to find a matching when there is no stable matching.

In particular, for every agent~$x$, our method extends the preference list of~$x$ by including every agent~$y$ who is not already in~$x$'s preference list but ``$k$-connected" to~$x$ (i.e.,~$x$ and~$y$ are connected a chain of~$k$ PFOAF---preferred friend of a friend---relations).

Let us introduce the relevant definitions and notation to make this idea more rigorous.

\subsection{Defining~$k$-acceptable pairs with respect to~$\prec_{x}$}


Let~$A_x^{+}$ be a set of {\em acceptable} agents to~$x \in A$, and $A_x^{-}$ be a set of {\em unwanted} agents by~$x \in A$. Note that~$A_x^{+} \cup A_x^{-} \subseteq A$. Let~$A^{-}$ be a set of unwanted pairs, \ie~$A^{-} = \{\{x,y\}: x \in A_y^{-},~y\in A\}$.
%
If an agent~$x$ states~$y$ as acceptable or 
unwanted, then we say that 
$y$ is {\em known} by~$x$.

Let~$K_x$ be a set of agents known by~$x$. We assume that~$A_x^{+} \cup A_x^{-} = K_x$.
We denote by~$K$ the set of pairs of agents who 
are known by each other, 
\ie~${K = \{\{x,y\}: x \in K_y,~y\in A\}}$. { Note that $K$ does not require that agents mutually know each other.}


A {\em~$k$-acceptability graph} is an undirected graph $G_A=(V,E)$, where 
every vertex $x \in V$ uniquely denotes an agent in~$A$, and every edge~$\{x,y\}\in E$ denotes pairs of agents that know each other and none of them is unwanted by the other: $\{x,y\} \in K \setminus A^{-}$. The curious reader may notice that the ``known by" relation is not symmetric but the graph $G_A$ is undirected (since it includes an undirected edge to denote one-sided known but non-unwanted pairs). We construct $G_A=(V,E)$ a bit more generously based on the following assumption: if a student $x$ is known by another student $y$, and neither student finds the other one unwanted, then the pair $\{x,y\} \in E$ regardless of $y$ is known by $x$. This assumption is reasonable and allows more possibilities for good-enough matchings. Indeed, a student $x$ might state someone $y$ in their preference list without being listed in return because the other student $y$ has not provided any preferences; in such cases, it is reasonable to assume that $x$ is not unwanted by $y$ and it is possible that $y$ might include $x$ in their list if they provide their preferences.


We say that agents~$x$ and~$y$ form a {\em \k-cp} if there exists a path of length~$k$ that connects $x$ and~$y$ in $G_A$.
Let us denote by $P^{k}_{G_A}$ the set of all \k-cp in~$G_A$, and, for simplicity, we will refer to it as $P^k$ throughout the remainder of the paper.

\begin{example}
Consider the running example. Let us denote a pair~$\{x,y\}$ of students in~$A$ by $xy$.
Suppose that~$b$ states that~$d$ is 
unwanted $A_{b}^{-} = \{d\}$.
Then,~$A^{-}=\{bd\}$.
The~$k$-acceptability graph $G_A=(V,E)$ is defined as follows:
$V= \{a,~b,~c,~d,~e\} $ and
$E= \{ae,~bc,~be,~de\} $.
According to the $k$-acceptability graph $G_A=(V,E)$, \k-cp for $k=1,2,3$ are defined as follows:
$P^1 = \{ae,~bc,~be,~de\}$, $P^2 = \{ad,~ab,~bd,~ce\}$, and $P^3 = \{ac,~cd\}$.
\end{example}

\subsection{Breaking ties in~$\prec_{x}^{'}$ by considering \k-cp}


Let~$I_x$ be a set of agents that are inferred as acceptable to~$x$ as roommates, \ie~$I_x = \{y: y \in \prec'_{x}\}$. If an agent~$x$ is indifferent between~$y$ and~$z$ in~$I_x$, then this tie is broken by considering \k-cp according to the following rule: 
if there exists a value~$k$ such that~$\{x,y\} \in P{^k}$, and there is no $l$ such that ~$\{x,z\} \in P{^l}$ for~$l\leq k$, then~$y \prec'_{x} z$.

\begin{example}
Consider the running example. According to~$\prec'_{c}$, student~$c$ is indifferent between student~$a$ and~$e$. We can break this tie by considering \k-cp:~$e \prec_{c}' a$ since there exist a value~$k=2$ such that~$ce \in P{^2}$ and~$ac \notin P{^2}$, but~$ac \in P{^3}$.
\end{example}

\subsection{Defining the~$k$-extended preference lists~$\prec_{x}^{k''}$}

Let~$O_{x}$ be a set 
of agents that may be acceptable to~$x$, \ie~$O_x = \{y: y \in A \backslash {(K_x} \cup I_x)\}$. Let~$\prec_{x}^{k}$ be a partial ordering of~$x$'s preferences over~$O_{x}$ where incomparability is transitive. 
For two agents~$y$ and~$z$ in~$O_x$,~$y \prec_{x}^{k} z$ if~$\{x,y\} \in P{^{i}}$ and~$\{x,z\} \notin P{^{i}}$, but~$\{x,z\} \in P{^{k}}$ for some~$0<i<k$. Note that for every~$y \in O_{x}$,~$y \in \prec_{x}^1$ if~$\{x,y\} \in P{^1}$.

Considering \k-cp, we define~$\prec_{x}^{k''} = \prec_{x}^{''} + \prec_{x}^{k}$ as a {\em~$k$-extended preference list}. Note that~$\prec_{x}^{k''}$ is the same as~$\prec_{x}^{''}$ when~$k=0$. Let us denote by $\prec^{k''}$ the collection of the~$k$-extended preference lists of each agent~$x \in A$.

\begin{example}
For the running example, Table~\ref{instance1-1} shows $k$-extended preference lists for each student, for $k=1$ and $k=2$. Students~$a$ and~$c$ may be acceptable to~$b$ by~$O_{b}= \{a,~c\}~$. Since~$bc \in P{^1}$, ${ab \notin P^1}$ and~${ab \in P^2}$,~$c \in \prec_{b}^1$ and~$c \prec_{b}^{2} a$.
\end{example}

\begin{table}[t]
\caption{A~$k$-Personalized-\srti instance, characterized by a set of students~$x$, sets~$A_x^{-}$ of unwanted students as roommates, and the $k$-extended lists~$\prec_{x}^{k''}$ of preferred students as roommates for $k=1$ and $k=2$. }
\label{instance1-1}
{\tablefont\begin{tabular}{@{\extracolsep{\fill}}cccccccc}
\topline
 &  \multicolumn{7}{c}{Unwanted sets and preference lists} \\
\cline{2-8}
Student~$x$  & $A_x^{-}$ & $\prec_{x}$ & $\prec_{x}^{'}$ & $\prec_{x}^{1}$  & $\prec_{x}^{1''}$ & $\prec_{x}^{2}$ & $\prec_{x}^{2''}$
\midline
$a$  & 	& $\langle e \rangle$ & $\langle b \rangle$ & & $\langle e,~b\rangle$ & $\langle d \rangle$ & $\langle e,~b,~d\rangle$ \\
$b$  & $\{d\}$  &	$\langle e \rangle$ &  &  $\langle c \rangle$ & $\langle e,~c \rangle$  & $\langle c,~a \rangle$ & $\langle e,~c,~a \rangle$ \\
$c$  & 	& $\langle b \rangle$   & $\langle e,~a \rangle$ & & $\langle b,~e,~a\rangle$  &  &  $\langle b,~e,~a\rangle$ \\
$d$  &  & $\langle b \rangle$  &  &  $ \langle e \rangle$ & $\langle b,~e\rangle$ & $\langle e,~a  \rangle$ & $\langle b,~e,~a\rangle$ \\
$e$  &  & $\langle d \rangle$  &  &  $ \langle \{a,b\} \rangle$ & $\langle d,\{a,b\} \rangle$ & $\langle \{a,b\},~c \rangle$ & $\langle d,\{a,b\},~c\rangle$
\botline
\end{tabular}}
\end{table}


A {\em $k$-Personalized-\srti instance} is an \srti instance characterized by a set $A$ of agents,
a set $A^-$ of pairs of agents (denoting which agents are forbidden for which), and a collection $\prec^{k''}$ of the~$k$-extended preference lists of each agent~$x \in A$.  A solution to a $k$-Personalized-\srti instance is defined in the following section, as ``$k$-stable matchings''.

\subsection{Defining~$k$-matchings with respect to~$\prec_{y}^{k''}$}


We say that~$x$ is {\em~$k$-acceptable} to~$y$ if~$x \in \prec_{y}^{k''}$. If an agent~$x$ is~$k$-acceptable to~$y$ and~$y$ is~$k$-acceptable to~$x$, then~$x$ and~$y$ are called {\em~$k$-mutually acceptable}.


A {\em $k$-matching} for a given~$k$-Personalized-\srti instance $(A,A^-,\prec^{k''})$, is a function~${M: A \mapsto A}$ such that, for all~$\{x,y\} \subseteq A\times A$ where~$x\in A_{y}{^{+}} \cup I_y \cup O_y$ and~$y \in  A_{x}{^{+}} \cup I_x \cup O_x$,~$M(x){=}y$ if and only if~$M(y){=}x$.

A pair~$\{x,y\}\subseteq A\times A~(x \neq y)$ is a {\em~$k$-blocking pair} with respect to~$M$ 
if
\begin{itemize}
	\item[i)]~$x$ and~$y$ are~$k$-mutually acceptable to each other,
    \item[ii)]~$x$ is single in~$M$, or~$y \prec_{x}^{k''} M(x)$, and
    \item[iii)]~$y$ is single in~$M$, or~$x \prec_{y}^{k''} M(y)$.
\end{itemize}

A~$k$-matching~${M}$ is called {\em~$k$-stable} if there is no~$k$-blocking pair with respect to~$M$.

\begin{example}
For the running example , we can find a~$1$-stable matching~$M_1 = \{a,bc,de\}$. This solution is also~$2$-stable.
\end{example}

%

{\em\srti with~$k$-connection ($k$-\srti)} is a special case of~$k$-Personalized-\srti $(A,\prec^{k''})$ where, for every agent~$x\in A, \prec_{x}^{''} = \prec_{x}$).

\vspace{2ex}
{\begin{proposition}\label{thm:complexity}
$k$-\srti (and thus~$k$-Personalized-\srti) is NP-complete.
\end{proposition}
}


\subsection{Observations about~$k$-Personalized-\srti}

{\it Observation 1:~$k$-stability is not monotone.} Consider the example shown in Table~\ref{instance2}. Consider~$I=(A,\prec)$ as an instance of \srti, and ${I'=(A,\prec^{''})}$ as an instance of Personalized-\srti. For both of them, there is no stable matching since there is no {student} who prefers each other. Then, we consider~${I''=(A,A^{-},\prec^{1''})}$ as an instance of~$1$-\srti. There are two~$1$-stable matchings:~$M_1 = \{ab,cd,e,f\}$, and~$M_2=\{af,bc,d,e\}$. If we further consider~$I'''={(A,A^{-},\prec^{2''})}$ as an instance of~$2$-\srti,~$M_1$ is not a~$2$-stable matching 
but~$M_2$ is a~$2$-stable matching.

\begin{table}[t]
\caption{A~$k$-Personalized-\srti instance, characterized by a set of students~$x$, sets~$A_x^{-}$ of unwanted students as roommates, and the $k$-extended lists~$\prec_{x}^{k''}$ of preferred students as roommates for $k=1$ and $k=2$.}
\label{instance2}
{\tablefont\begin{tabular}{@{\extracolsep{\fill}}cccccccc}
\topline
 & \multicolumn{7}{c}{Unwanted sets and preference lists} \\
\cline{2-8}
Student~$x$ & $A_x^{-}$ & $\prec_{x}$  & $\prec_{x}^{'}$       &$\prec_{x}^{1}$           & $\prec_{x}^{1''}$             & $\prec_{x}^{2}$            & $\prec_{x}^{2''}$
\midline
$a$         & 	 & $\langle b \rangle$  & $\langle d \rangle$   & $\langle f \rangle$      & $\langle b,d,f \rangle$       &$\langle f, c\rangle$       &  $\langle b,d,f,c \rangle$  \\

$b$         & 	 &						 & $\langle f,e \rangle$ & $\langle \{a,c\}\rangle$ & $\langle f,e,\{a,c\} \rangle$ &$\langle \{a,c\}, d\rangle$ & $\langle f,e,\{a,c\},d \rangle$   \\

$c$     &$\{e\}$ & $\langle b \rangle$  &                       & $\langle d \rangle$      & $\langle b,d \rangle$         &$\langle d, a\rangle$       &  $\langle b,d,a \rangle$\\

$d$         &    & $\langle c \rangle$  & $\langle b \rangle$   &                          & $\langle c,b \rangle$         &  							 &$\langle c,b \rangle$  \\
$e$         & 	 & $\langle c \rangle$  & $\langle a \rangle$   &                          &  $\langle c,a \rangle$        &  							 &$\langle c,a \rangle$  \\
$f$         & 	 & $\langle a \rangle$  & $\langle c \rangle$   &                          & $\langle a,c \rangle$         & $\langle b  \rangle$        &$\langle a,c,b \rangle$
\botline
\end{tabular}}
\end{table}

\begin{table}[t]
\caption{A~$k$-Personalized-\srti instance, characterized by a set of students~$x$, sets~$A_x^{-}$ of unwanted students as roommates, and the $k$-extended lists~$\prec_{x}^{1''}$ of preferred students as roommates for $k=1$.} \label{instance3}
\centering
{\tablefont\begin{tabular}{@{\extracolsep{\fill}}cccccc}
\topline
& \multicolumn{5}{c}{Unwanted sets and preference lists} \\
\cline{2-6}

Student~$x$  & $A_x^{-}$ & $\prec_{x}$          & $\prec_{x}^{'}$ & $\prec_{x}^1$ & $\prec_{x}^{1''}$
\midline
$a$          & 		& $\langle b \rangle$     &      &                             & $\langle b \rangle$  \\

$b$          & 	    & $\langle e,~a \rangle$  &      &   $\langle \{c,d\} \rangle$ & $\langle e,a,\{c,d\} \rangle$  \\

$c$          & 	    & $\langle b,~d \rangle$  &      &   $\langle e \rangle$       &  $\langle b,d,e \rangle$   \\

$d$          & 	    & $\langle b,~c \rangle$  &      &                             & $\langle b,c \rangle$ \\

$e$          & 		&			              &      &   $\langle b \rangle$        & $\langle b \rangle$ \\
\hline \hline
\end{tabular}}
\end{table}

{\it Observation 2: A stable matching may not be a~$k$-stable matching.} Consider the example shown in Table~\ref{instance3}. Consider the \srti instance~$I=(A,\prec)$, and  Personalized-\srti instance~${I'=(A,\prec^{''})}$. For both of them, there is a stable matching:~${M=\{ab,cd,e\}}$. Then, for the~$1$-\srti instance~${I''=(A,A^{-},\prec^{1''})}$,~$M$ is not a~$1$-stable matching since~$be$ blocks the matching~$M$. Hence,~$M_1=\{be,cd,a\}$ is a~$1$-stable matching.


\section{Solving~$k$-Personalized-\srti using ASP}

We formalize the input 
$I=(A,{ A^{-}},\prec^{k''})$
of a~$k$-Personalized-\srti instance in ASP by a set~$F_I$ of facts using atoms of the forms~$\ii{agent}(x)$ (``$x$ is an agent in~$A$''), 
{~$\ii{unwanted(x,y)}$} (``agent~$x$ states that agent~$y$ is {unwanted}, i.e.,~$y \in {A_x^{-}}$''), and~$\ii{kPrefer}(x,y,z)$ (``agent~$x$ prefers agent~$y$ to agent~$z$ by degree~$k$, i.e.,~$y~\prec_{x}^{k''}~z$'').

\paragraph{\it Preferences of each agent in~$\prec_{x}$ and~$\prec_{x}^{'}$.}
We use atoms of the forms~$\ii{prefer2}(x,y,z)$ (``agent~$x$ prefers agent~$y$ to agent~$z$, i.e.,~$y \prec_{x} z$''),~$\ii{iprefer}(x,y,z)$ (``agent~$x$ prefers agent~$y$ to agent~$z$ in terms of the criteria-based personalized preference lists $y \prec_{x}^{'} z$''), and~$\ii{tie}(x,y,z)$ (``agent~$x$ is indifferent between agent~$y$ and agent~$z$'').

Recall that, based on the stated preferences of agents, we define acceptable pairs (cf. rules~\ref{eq:P-accept}). Similarly, based on the inferred preferences by~$\prec_{x}^{'}$, we define inferred acceptable pairs:

\vspace{1ex}
$
\ba l
\ii{inferred}(x,y) \lar \ii{iprefer}(x,y,\_), \no\ \ii{prefer2}(x,y,\_). \\
\ii{inferred}(x,y) \lar \ii{iprefer}(x,\_,y).
\ea
$

\paragraph{\it  Defining~$k$-acceptability graphs.}
Based on the stated
{unwanted} agents, we define unacceptable pairs:

\vspace{1ex}
$\ba l
\ii{not\_accept}(x,y) \lar 1\{\ii{unwanted}(x,y);\ii{unwanted}(y,x)\} \qquad (x\neq y).
\ea$

\vspace{1ex}
\noindent Considering also the acceptable pairs, 
we define pairs of agents who know each other:

\vspace{1ex}
$
\ba l
\ii{know}(x,y) \lar 1\{\ii{accept}(x,y);\ii{accept}(y,x);\ii{not\_accept}(x,y)\} \qquad (x\neq y).
\ea
$

\vspace{1ex}
\noindent After we define the edges in the~$k$-acceptability graph:

\vspace{1ex}
$\ba l
\ii{edge}(x,y) \lar \ii{know}(x,y),~\no\ \ii{not\_accept}(x,y) \qquad (x\neq y).
\ea$

\vspace{1ex}
\noindent we define \k-cp recursively:

\vspace{1ex}
$\ba l
\ii{connected}(x,y,1) \lar \ii{edge}(x,y) \qquad (x\neq y, k>0). \\
\ii{connected}(x,y,i+1) \lar \ii{connected}(x,z,i),~\ii{edge}(z,y) \qquad (x\neq y, i<k).
\ea$

\paragraph{\it Breaking ties.}
Considering \k-cp, ties in~$\prec_{x}^{'}$ are broken:

\vspace{1ex}
$\ba l
\ii{iprefer}(x,y,z) \lar \ii{tie}(x,y,z),~\ii{connected}(x,y,i),~\no\ \ii{connected}(x,z,i). \\
\ea$

\paragraph{\it Defining the~$k$-extended preference lists~$\prec_{x}^{k''}$.}
Based on~$\prec_{x}^{k}$, we define the concept of may be acceptable agents~$y$ 
($y \in  A \backslash ( K_x \cup I_x) {\text{ and }} \{x,y\} \in P{^k}$) with degree~$k$:

\vspace{1ex}
$\ba l
\ii{may\_accept}(x,y,i) \lar \no\ \ii{accept}(x,y),~\no\ \ii{inferred}(x,y), \\
\qquad \qquad \qquad \qquad \qquad \no\ \ii{unwanted}(x,y), ~\ii{connected}(x,y,i).
\ea$

\paragraph{\it Extending the preferences to define~$k$-matching.}
Based on the preferences of each agent in~$\prec_{x}$ and~$\prec_{x}^{'}$, we define extended preferences in~$\prec_{x}^{''}$:

\vspace{1ex}
$\ba l
\ii{ePrefer}(x,y,z) \lar \ii{prefer2}(x,y,z). \\
\ii{ePrefer}(x,y,z) \lar \ii{iprefer}(x,y,z). \\
\ii{ePrefer}(x,y,z) \lar \ii{prefer2}(x,\_,y),~\ii{iprefer}(x,z,\_). \\ 
\ii{ePrefer}(x,z,x) \lar \ii{iprefer}(x,\_,z),~\no\ \ii{prefer2}(x,z,\_) \\
\ea$

\vspace{1ex}
Based on the preferences of each agent in~$\prec_{x}^{''}$, we extend preferences by degree~$k$ recursively:

\vspace{1ex}
$\ba l
\ii{prefer}(x,z,x,j) \lar \ii{may\_accept}(x,z,j) \quad (x\neq z,~j<k+1). \\
\ii{prefer}(x,y,z,j) \lar \ii{ePrefer}(x,y,\_),~\ii{may\_accept}(x,z,j). \\
\ii{prefer}(x,y,z,j) \lar \ii{prefer}(x,y,\_, i),~\no\ \ii{ePrefer}(x,z,\_), \ii{may\_accept}(x,y,i), \\
\qquad \qquad \ii{may\_accept}(x,z,j), \no\ \ii{may\_accept}(x,z,i) \quad (x\neq y,~ y\neq z,~i<j). \\
\ii{prefer}(x,y,z,j) \lar \ii{prefer}(x,\_,y, i),~\no\ \ii{ePrefer}(x,z,\_), \ii{may\_accept}(x,y,i), \\
\qquad \qquad \ii{may\_accept}(x,z,j), \no\ \ii{may\_accept}(x,z,i) \quad (x\neq y,~ y\neq z,~i<j).
\ea$

\vspace{1ex}
Finally, we define the preferences of each agent in~$\prec_{x}^{k''}$:

$\ba l
\ii{kPrefer}(x,y,z) \lar \ii{ePrefer}(x,y,z) \qquad (x\neq y). \\
\ii{kPrefer}(x,y,z) \lar \ii{prefer}(x,y,z,i) \qquad (x\neq y,~i<k+1). \\
\ea$

\paragraph{\it~$k$-matching.}
A $k$-Personalized-\srti instance is characterized by 
$(A,{A^{-}},\prec^{k''})$ where~$A$ is a finite set of agent, {~$A^{-}$} is a finite set of {unwanted} pairs, and~$\prec^{k''}$ is the collection of the~$k$-extended preference list of each agent~$x \in A$. To solve a~$k$-Personalized-\srti instance, i.e., to compute a $k$-matching, we utilize \srtiasp~\citep{erdem2020} (as briefly described in 
Section \ref{sec:srti})
using the atoms of the form~$\ii{kPrefer}(x,y,z)$ instead of~$\ii{prefer2}(x,y,z)$.

\section{Experimental Evaluations}


In our earlier work~\citep{fidan2021}, we have experimentally evaluated our ASP-based method for Personalized-\srti, to better understand its scalability, usefulness and applicability, by utilizing the domain-specific knowledge inferred from the agents' habits and habitual preferences.

In this study, we investigate the usefulness and the applicability of utilizing the knowledge about the networks of the agents' preferred friends. We design and perform two types of experiments to study the following questions:

\begin{itemize}
	\item[Q1] When there is no stable matching to a given \srti instance or a good-enough matching to its corresponding Personalized-\srti instance that also considers the agents' habitual preferences, can we compute good-enough matchings considering also the networks of the agents' preferred friends?
    \item[Q2] Does it make sense to compute good-enough matchings by considering the networks of the agents' preferred friends?
\end{itemize}

In the first set of experiments (addressing Q1), we evaluate the usefulness of the additional knowledge about the networks of the agents' preferred friends, from the computational perspectives. We randomly generate instances and evaluate the results with respect to objective measures (i.e., how many more instances can be solved with this additional knowledge?).

In the second set of experiments (addressing Q2), we evaluate the usefulness of the additional knowledge from the perspective of the users. We perform polls, surveys and interviews with students to obtain their feedback, and use subjective measures to evaluate the results.

The full encoding and benchmark instances are made available on the following GitHub page:~\url{https://github.com/mugefidan/k-SRTI.git}.

\subsection{Experimental Evaluation with Objective Measures}


\paragraph{Benchmark instances.}
First we randomly generate \srti instances using Prosser's software~\citep{sricp2020} based on Merten's idea~(\citeyear{Mertens2005}): (1) we generate a random graph ensemble~$G(n,p)$ according to the Erd\"{o}s-R\'{e}nyi model~(\citeyear{Erdos60}), where~$n$ is the required number of agents and~$p$ is the edge probability (i.e., each pair of vertices is connected independently with probability~$p$); (2) since the edges characterize the acceptability relations, we generate a random permutation of each agent's acceptable partners to provide the preference lists.

Then, motivated by the real world applications, (3) in each \srti instance, we shorten the given preference lists so as to contain the first 5 choices for each agent. Then, (4) we populate each instance to a set of Personalized-\srti instances, by extending these lists by including the choices inferred from the habitual preferences of agents:  we consider 2 and 5 habitual criteria, randomly generate the importance of each criterion, and randomly generate the habitual preferences of 20\% and 50\% of agents.

For each instance generated with the method above, we define its  {\em completeness degree}  (c.d.) as the probability of an agent occurring in a preference list (i.e., the edge probability). We define the {\em mutual acceptability rate} (m.a.p.) of an instance as the ratio of the number of mutually acceptable pairs to the number of all possible pairs. 
Consider the example shown in Table~\ref{instance1-1}. 
For a given $2$-Personalized-\srti instance, there are~$14$ mutually acceptable pairs out of 16 possible pairs. There are two pairs, such as~$\{a,c\}$ and~$\{b,d\}$, that are not mutually acceptable. Hence, the mutually acceptability rate is~$\frac{14}{16}$, which simplifies to~$0.875$.

We generate two sets of benchmarks, with respect to their m.a.p.: HMA instances have high m.a.p. of 75\% or more, while LMA instances have lower m.a.p..
%


For HMA instances, first, as described in steps (1) and (2) above, we generate \srti instances with 40, 60, 80, 100, 150 and 200 agents, with c.d. of 0.075, 0.05, 0.0375, 0.03, 0.02, and 0.015, respectively. Such low c.d. allow us to generate \srti instances with preference lists with approximate length 3.
%

In our earlier study~\citep{erdem2020} that investigates the scalability of our approach to solve \srti problems, we have already generated benchmark instances using Prosser's software, as described in steps (1) and (2) above.
For LMA instances, we start with those instances with 40, 60, 80, 100, 150 and 200 agents, and with c.d. 0.25.

Then, in each benchmark set, for each number of criteria and the percentage of agents who specify their habitual preferences, we populate each \srti instance by generating 20 Personalized-\srti instances, as described in steps (3) and (4) above.
%

The characteristics of each set of benchmarks are summarized in Tables~\ref{tab:highMutually} and~\ref{tab:lowMutually}. Consider, for instance, the first row of Table~\ref{tab:highMutually} that describes  average values for 20 HMA instances with 40 agents. The \srti instances (generated after steps (1), (2) and (3)) have an average c.d. of 0.073 and an average m.a.p. of 0.965.  After step (4) where we add knowledge about the habitual preferences of 50\% of agents by considering 5 criteria, the average c.d. increases to 0.207 while the average m.a.p. decreases to 0.757.

\begin{table}[t]
\caption{HMA Instances: Average completeness degree (c.d.) and average percentage of mutually acceptability rate (m.a.p.).}
\label{tab:highMutually}
\centering
\begin{tabular}{c|cc|cccc|cccc}
\topline
\multirow{3}{*}{\#agents} &
  \multicolumn{2}{c|}{\multirow{2}{*}{SRTI}} &
  \multicolumn{4}{c|}{2 criteria} &
  \multicolumn{4}{c}{5 criteria}  \\ \cline{4-11}
 &
  \multicolumn{2}{c|}{} &
  \multicolumn{2}{c|}{20\% response} &
  \multicolumn{2}{c|}{50\% response} &
  \multicolumn{2}{c|}{20\% response} &
  \multicolumn{2}{c}{50\% response} \\ \cline{2-11}
 &
  \multicolumn{1}{c}{c.d.} &
  \multicolumn{1}{c|}{m.a.p.} &
  \multicolumn{1}{c}{c.d.} &
    \multicolumn{1}{c|}{m.a.p.}  &
  \multicolumn{1}{c}{c.d.} &
   \multicolumn{1}{c|}{m.a.p.}  &
  \multicolumn{1}{c}{c.d.} &
    \multicolumn{1}{c|}{m.a.p.} &
  \multicolumn{1}{c}{c.d.} &
    \multicolumn{1}{c}{m.a.p.}  \\ \cline{1-11}
40 &
  \multicolumn{1}{c|}{0.073} &
  0.965 &
  \multicolumn{1}{c|}{0.085} &
\multicolumn{1}{c|}{0.915} &
\multicolumn{1}{c|}{0.092} &
\multicolumn{1}{c|}{0.837} &
\multicolumn{1}{c|}{0.1} &
\multicolumn{1}{c|}{0.911} &
\multicolumn{1}{c|}{0.207} &
\multicolumn{1}{c}{0.757} \\ \cline{1-11}
60 &
  \multicolumn{1}{c|}{0.048} &
  0.96 &
  \multicolumn{1}{c|}{0.054} &
\multicolumn{1}{c|}{0.9} &
\multicolumn{1}{c|}{0.058} &
\multicolumn{1}{c|}{0.82} &
\multicolumn{1}{c|}{0.073} &
\multicolumn{1}{c|}{0.864} &
\multicolumn{1}{c|}{0.132} &
\multicolumn{1}{c}{0.736} \\ \cline{1-11}
80 &
  \multicolumn{1}{c|}{0.037} &
  0.952 &
 \multicolumn{1}{c|}{0.041} &
\multicolumn{1}{c|}{0.892} &
\multicolumn{1}{c|}{0.044} &
\multicolumn{1}{c|}{0.824} &
\multicolumn{1}{c|}{0.062} &
\multicolumn{1}{c|}{0.829} &
\multicolumn{1}{c|}{0.101} &
\multicolumn{1}{c}{0.727} \\ \cline{1-11}
100 &
  \multicolumn{1}{c|}{0.03} &
  0.954 &
 \multicolumn{1}{c|}{0.033} &
\multicolumn{1}{c|}{0.895} &
\multicolumn{1}{c|}{0.035} &
\multicolumn{1}{c|}{0.82} &
\multicolumn{1}{c|}{0.049} &
\multicolumn{1}{c|}{0.835} &
\multicolumn{1}{c|}{0.075} &
\multicolumn{1}{c}{0.732} \\ \cline{1-11}
150 &
  \multicolumn{1}{c|}{0.02} &
  0.957 &
  \multicolumn{1}{c|}{0.021} &
\multicolumn{1}{c|}{0.898} &
\multicolumn{1}{c|}{0.023} &
\multicolumn{1}{c|}{0.821} &
\multicolumn{1}{c|}{0.032} &
\multicolumn{1}{c|}{0.821} &
\multicolumn{1}{c|}{0.044} &
\multicolumn{1}{c}{0.738}  \\ \cline{1-11}
200 &
  \multicolumn{1}{c|}{0.015} &
  0.955 &
  \multicolumn{1}{c|}{0.016} &
\multicolumn{1}{c|}{0.895} &
\multicolumn{1}{c|}{0.017} &
\multicolumn{1}{c|}{0.821} &
\multicolumn{1}{c|}{0.024} &
\multicolumn{1}{c|}{0.81} &
\multicolumn{1}{c|}{0.029} &
\multicolumn{1}{c}{0.752} \\ \hline \hline
\end{tabular}%
\end{table}

\begin{table}[t]
\caption{LMA Instances: Average completeness degree (c.d.) and average percentage of mutually acceptability rate (m.a.p.).} \label{tab:lowMutually}
\centering
\resizebox{\textwidth}{!}{%
\begin{tabular}{c|cc|cccc|cccc}
\hline \hline
\multirow{3}{*}{\#agents} &
  \multicolumn{2}{c|}{\multirow{2}{*}{SRTI}} &
  \multicolumn{4}{c|}{2 criteria} &
  \multicolumn{4}{c}{5 criteria}  \\ \cline{4-11}
 &
  \multicolumn{2}{c|}{} &
  \multicolumn{2}{c|}{20\% response} &
  \multicolumn{2}{c|}{50\% response} &
  \multicolumn{2}{c|}{20\% response} &
  \multicolumn{2}{c}{50\% response} \\ \cline{2-11}
 &
  \multicolumn{1}{c}{c.d.} &
  \multicolumn{1}{c|}{m.a.p.} &
  \multicolumn{1}{c}{c.d.} &
    \multicolumn{1}{c|}{m.a.p.}  &
  \multicolumn{1}{c}{c.d.} &
   \multicolumn{1}{c|}{m.a.p.}  &
  \multicolumn{1}{c}{c.d.} &
    \multicolumn{1}{c|}{m.a.p.} &
  \multicolumn{1}{c}{c.d.} &
    \multicolumn{1}{c}{m.a.p.}  \\ \cline{1-11}
40 &
  \multicolumn{1}{c|}{0.127} &
  0.52 &
 \multicolumn{1}{c|}{0.138} &
\multicolumn{1}{c|}{0.532} &
\multicolumn{1}{c|}{0.157} &
\multicolumn{1}{c|}{0.563} &
\multicolumn{1}{c|}{0.156} &
\multicolumn{1}{c|}{0.571} &
\multicolumn{1}{c|}{0.241} &
\multicolumn{1}{c}{0.624} \\  \cline{1-11}
60 &
  \multicolumn{1}{c|}{0.085} &
  0.346 &
  \multicolumn{1}{c|}{0.092} &
\multicolumn{1}{c|}{0.366} &
\multicolumn{1}{c|}{0.102} &
\multicolumn{1}{c|}{0.421} &
\multicolumn{1}{c|}{0.114} &
\multicolumn{1}{c|}{0.45} &
\multicolumn{1}{c|}{0.183} &
\multicolumn{1}{c}{0.544} \\  \cline{1-11}
80 &
  \multicolumn{1}{c|}{0.063} &
  0.259 &
  \multicolumn{1}{c|}{0.068} &
\multicolumn{1}{c|}{0.287} &
\multicolumn{1}{c|}{0.076} &
\multicolumn{1}{c|}{0.349} &
\multicolumn{1}{c|}{0.086} &
\multicolumn{1}{c|}{0.382} &
\multicolumn{1}{c|}{0.135} &
\multicolumn{1}{c}{0.494} \\  \cline{1-11}
100 &
  \multicolumn{1}{c|}{0.051} &
  0.205 &
  \multicolumn{1}{c|}{0.054} &
\multicolumn{1}{c|}{0.236} &
\multicolumn{1}{c|}{0.061} &
\multicolumn{1}{c|}{0.319} &
\multicolumn{1}{c|}{0.072} &
\multicolumn{1}{c|}{0.35} &
\multicolumn{1}{c|}{0.105} &
\multicolumn{1}{c}{0.461} \\  \cline{1-11}
150 &
  \multicolumn{1}{c|}{0.034} &
  0.134 &
  \multicolumn{1}{c|}{0.036} &
\multicolumn{1}{c|}{0.168} &
\multicolumn{1}{c|}{0.041} &
\multicolumn{1}{c|}{0.268} &
\multicolumn{1}{c|}{0.047} &
\multicolumn{1}{c|}{0.29} &
\multicolumn{1}{c|}{0.065} &
\multicolumn{1}{c}{0.411} \\  \cline{1-11}
200 &
  \multicolumn{1}{c|}{0.025} &
  0.102 &
 \multicolumn{1}{c|}{0.027} &
\multicolumn{1}{c|}{0.135} &
\multicolumn{1}{c|}{0.031} &
\multicolumn{1}{c|}{0.238} &
\multicolumn{1}{c|}{0.036} &
\multicolumn{1}{c|}{0.267} &
\multicolumn{1}{c|}{0.045} &
\multicolumn{1}{c}{0.377} \\ \hline \hline
\end{tabular}%
}
\end{table}

\paragraph{Experimental results.}
We have evaluated our method over these two sets of benchmarks to understand the usefulness of adding knowledge about agents' networks of preferred friends. 


\begin{figure}[h!]
    \centering\includegraphics[width=0.55\columnwidth]{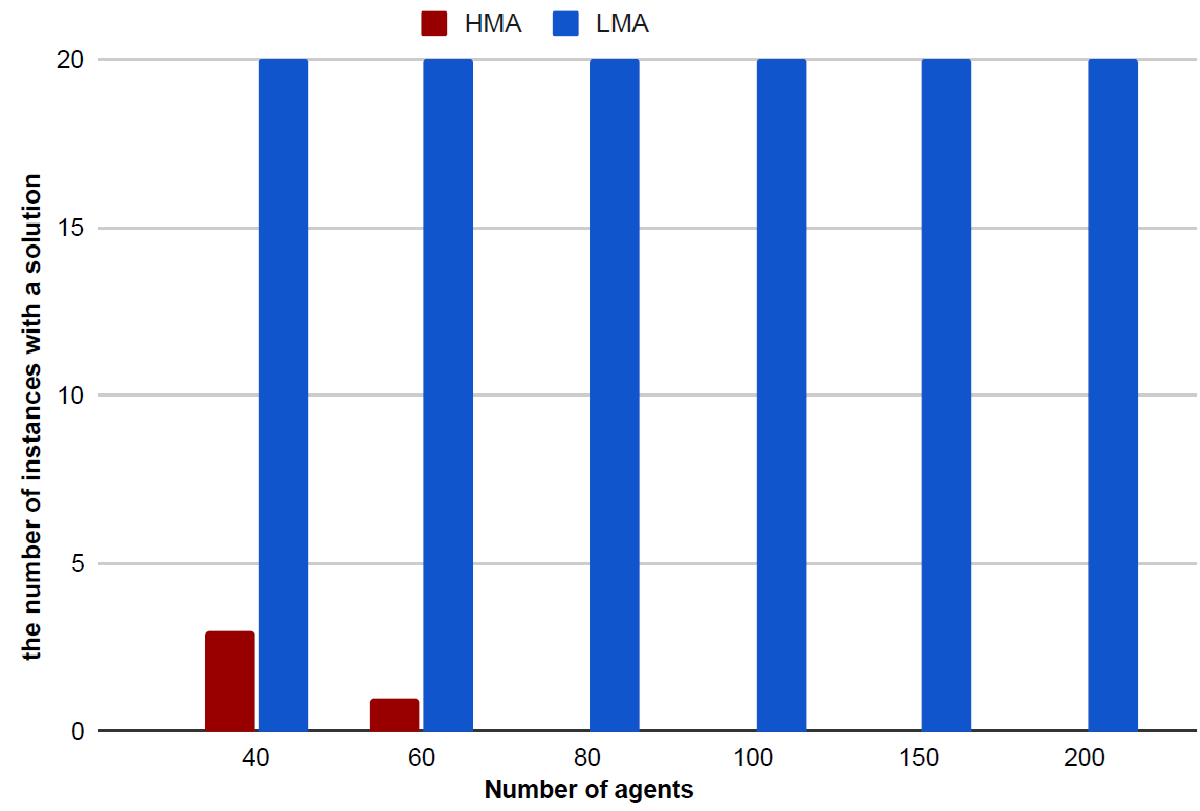}
    \caption{Usefulness of~$2$-\srti over \srti.}
\label{fig:resultsplot2}
\end{figure}

When there is no stable matching to the seed \srti instances, Figure~\ref{fig:resultsplot2} shows the computational benefit of extending the preference lists by only suitable candidates obtained from networks of preferred friends, considering~${2}$-connected pairs (i.e., preferred friend of a preferred friend): while only some {\em{HMA}} instances have good-enough matchings, almost all {\em{LMA}} instances have good-enough matchings. 
This {result} is expected because, as the preferences lists are extended, we have a higher number of mutually acceptable pairs, and this allows exploring more number of potential matchings, and thus increasing the chances of finding a good-enough solution.

\begin{figure}[h!]
\centering
\begin{tabular}{c}
  \includegraphics[valign=c,width=0.75\columnwidth]{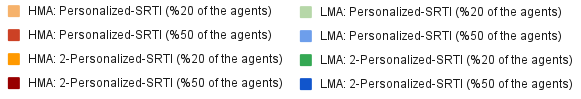} \\
  \includegraphics[valign=c,width=0.75\columnwidth]{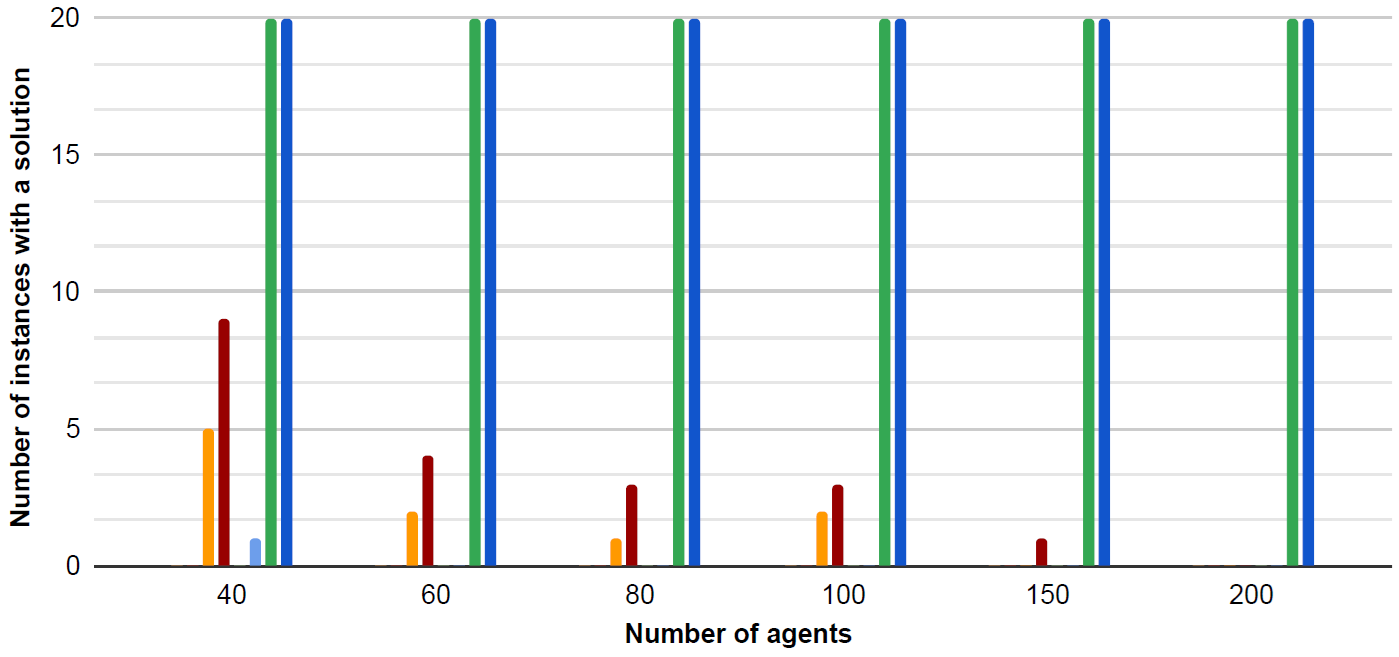} \\
  \includegraphics[valign=c,width=0.75\columnwidth]{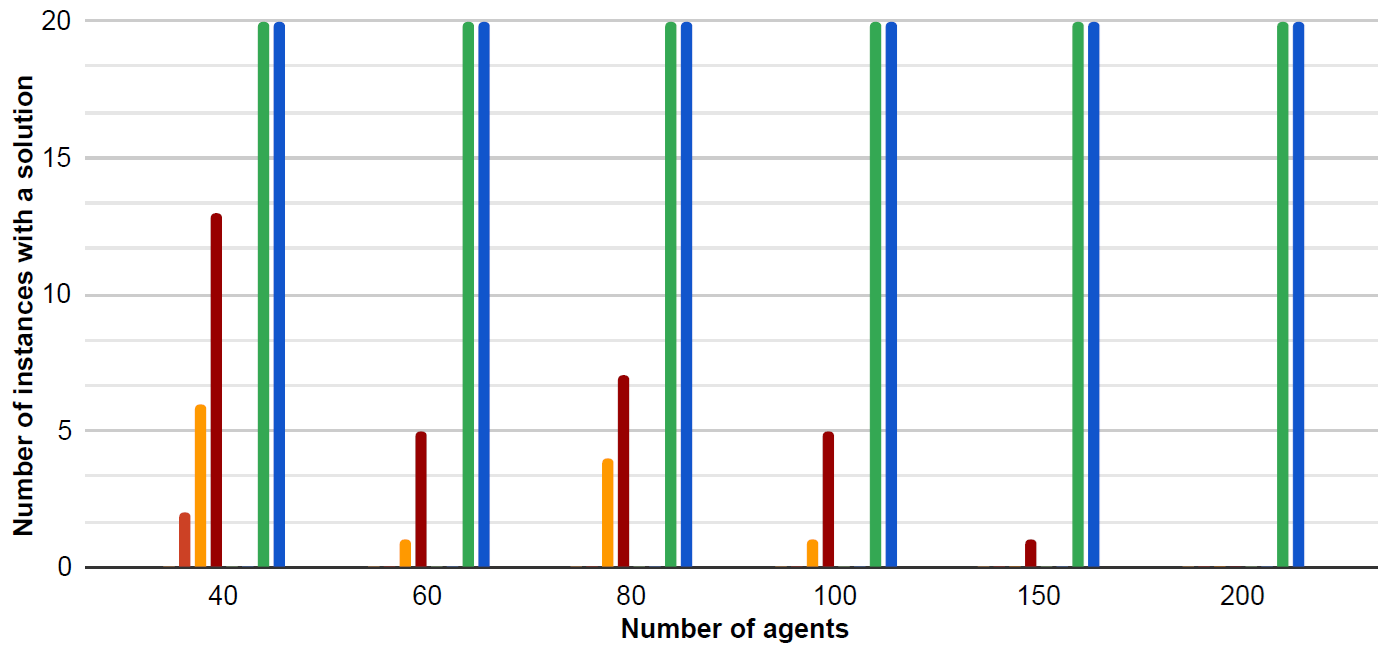} \\
\end{tabular}
\caption{Usefulness of~$2$-\srti over Personalized-\srti: (top) 2 criteria and (bottom) 5 criteria}
\label{all-srti}
\end{figure}

When there is no stable matching to the seed \srti instances and no good-enough solution after extending the preference lists considering only the agents' habits and habitual preferences, Figure~{\ref{all-srti}} shows that extending the preference lists by suitable candidates obtained from networks of preferred friends is still useful.

\begin{figure}[h!]
\centering
\begin{tabular}{cc}
  \includegraphics[width=0.47\textwidth]{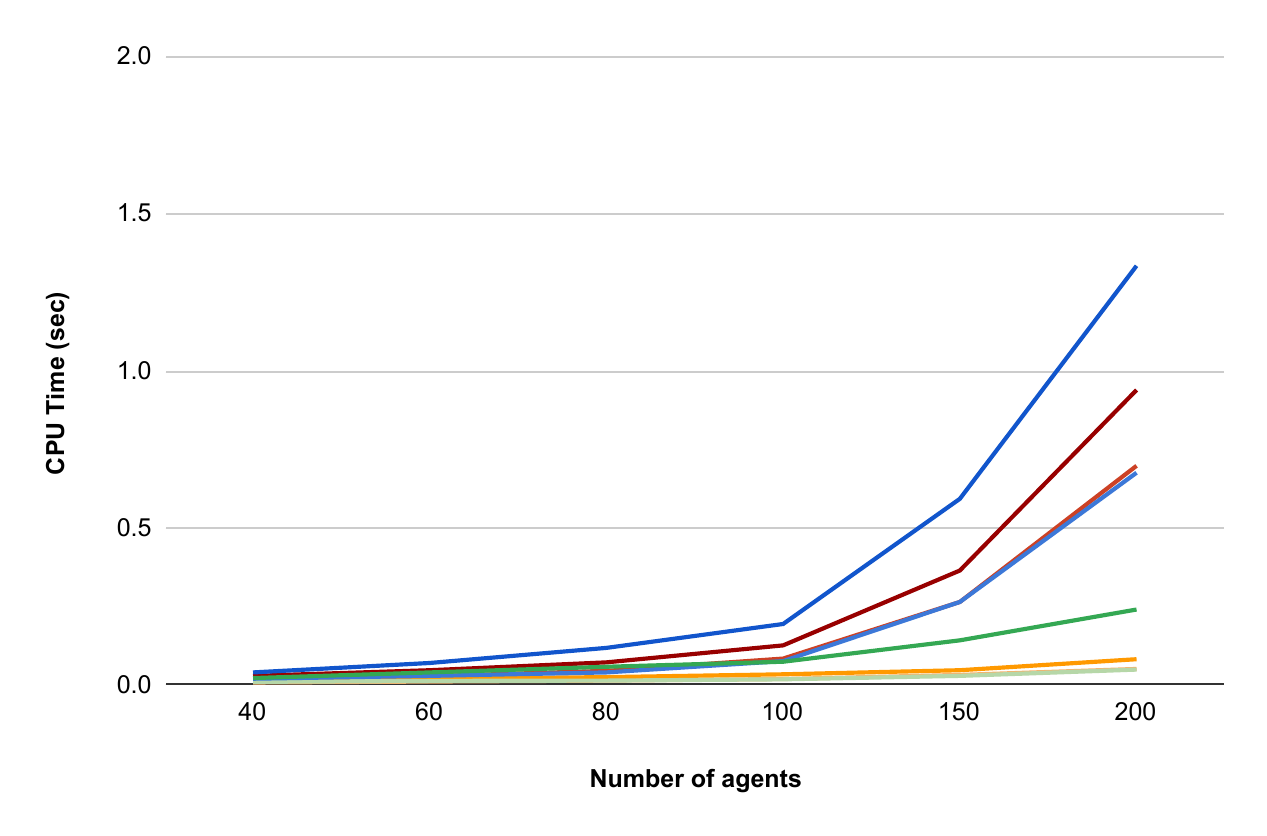} &
  \includegraphics[width=0.47\textwidth]{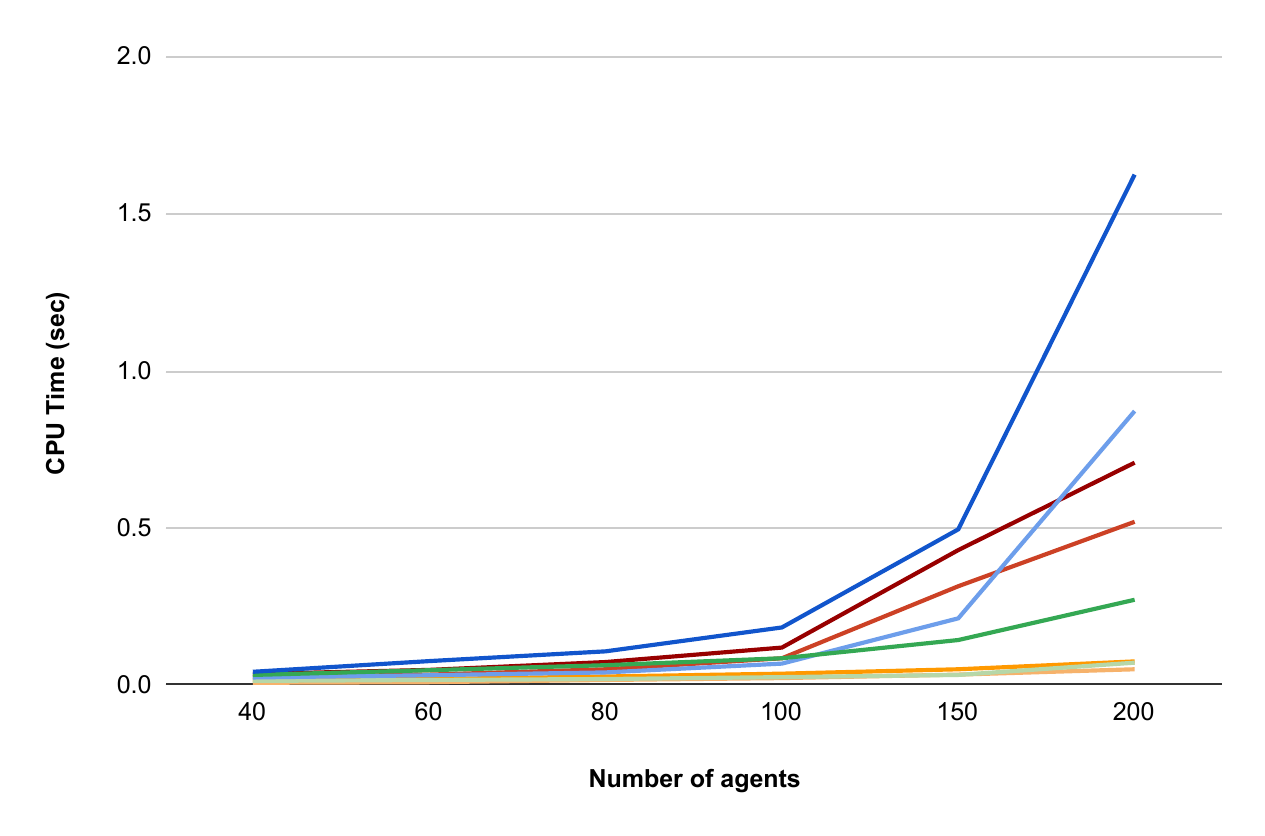} \\
\end{tabular}
\caption{Scalability of SRTI-ASP in computation time, for~$2$-\srti and Personalized-\srti: \quad (left) 2 criteria and (right) 5 criteria. Legend as in~Figure~\ref{all-srti}.}
\label{all-srti-cpu}
\end{figure}

In these experiments, the scalability of \srtiasp for $k$-\srti is also observed (Figure~\ref{all-srti-cpu}): computing good-enough matchings usually takes a few seconds  on a machine with Intel Xeon(R) W-2155 3.30GHz CPU and 32GB RAM.

%

\subsection{Experimental Evaluation with Subjective Measures}


We evaluate the usefulness of the additional knowledge from the perspective of the users. We design and perform poll survey and interviews with students 
to obtain their feedback, and use subjective measures to evaluate the results.

\paragraph{Online poll survey with students.} The objective is to find out whether it makes sense to match roommates considering the networks of preferred friends. It includes one question to learn whether the students prefer to roommate with someone from the preference lists of their preferred friends, when the dormitory system cannot match them with any of their friends on their preference lists. The survey, shown in~\ref{poll-survey}, is conducted online with 101 students at Sabanci University. In the end, 76 students prefer option of extending their preference lists with their preferred friends' preferences (i.e., option (a) in the poll).

\paragraph{Interviews with students.} The objective is to get more detailed information about their poll answers, and learn the underlying reasons of students' answers. Face-to-face interviews are conducted with {20 volunteers (8 female, and 12 male)} with diverse academic backgrounds. Each interview session is done one-to-one, taking an average of~$12-15$ minutes. Before the interview, each volunteer is informed about the objective and the process of the interview, and their permission is requested for recording the interviews. In the interview, the following questions are asked in an appropriate order based on their answers: ``Why did you choose this option at the poll?", ``Can you say a little more about \dots?", ``Did you have such an experience?", ``What if \dots", ``Why do you think \dots?".
To provide concrete examples to illustrate the kinds of questions asked and the kinds of responses gathered, two interview transcripts are included in Appendix A.

Among the volunteers, 4 students have experienced to be roommates with a friend of their friend, positively.
At the beginning of the interview, {14} volunteers have chosen option (a) of the poll, i.e., to be roommates with someone from the preference lists of their preferred friends. During the interview, {4} of the volunteers who chose option~(b) of the poll, i.e., to be roommates with anyone, have changed their decision. According to their final decisions, { 18} volunteers choose option (a).

Table~\ref{tab:reasonA} shows the volunteers' reasons for choosing option (a) of the poll, i.e., to be roommates with someone from the preference lists of their preferred friends, and how many of them give the same reason. For instance, most of the volunteers (16 out of 18) indicate that they trust the person their friend prefers to be roommate with, instead of someone they do not know anything about.

Table~\ref{tab:reasonB} displays the volunteers' reasons for choosing option~(b) of the poll, i.e., to be roommates with anyone, and how many of them give this reason. There are { 6 such students, and 4} of them have changed their decision during the interview. For instance, one volunteer have said ``I may have some friends, who have friends I do not like, in my preference list.". As a follow-up question, we have asked ``What if the dormitory system allows you to specify also the students you do not want to be matched with as roommates?", and the volunteer changed their preference from option~(b) to option~(a).

\begin{table}[t]
\caption{Out of 20 students, 14 students have chosen option (a), i.e., they prefer as roommates someone from the preference lists of their preferred friends, compared to anyone.  The reasons they have provided for this decision, the number of students who have given the same reason for their decisions before the interview and changed their minds during/after the interview are also provided. }
\label{tab:reasonA}
\centering
{\tablefont\begin{tabular}{@{\extracolsep{\fill}}lcc}
\topline
 & \multicolumn{2}{c}{\#Students}  \\
\cline{2-3}
Reasons for choosing option (a) & before  & during/after  \\
 & interview (14) & interview (18)
\midline
comfortable/trustful        &                   13 & 16    \\ \cline{1-3}
get along well   
(room environment)         &    8 & 10     \\ \cline{1-3}
get along well  
(personal characteristics) &    8 & 10     \\ \cline{1-3}
similarity 
(scholarship status, department/class)     &    5 & 5     \\ \cline{1-3}
sincerity, toleration                      &    4 & 5         \\ \cline{1-3}
mutual friend mediation 
(in case of disagreement)    &    3 & 4     \\ \hline	\hline	
\end{tabular}}
\end{table}

\paragraph*{Experimental results: quantitative comparisons.}
Most of the students (75.2\%) who participated in the online poll survey
have chosen option (a) at the poll survey. This result illustrates that it makes sense to consider the preferred friends of preferred friends while assigning roommates, and extending the preference lists with suitable candidates accordingly.
On the other hand,
only {10\%} of the volunteers in the interviews have chosen option~(b).  


\paragraph*{Experimental results: qualitative comparisons.}
The reasons for choosing option (a), obtained from the students
significantly coincide. For instance,
getting assigned to a roommate from the preference lists of their preferred friends is preferable (compared to random assignment) because they would be more comfortable and get along well with each other. These results also indicate the usefulness of our knowledge-based approach, from the users' point of view.



\begin{table}[t]
\caption{Out of 20 students, 6 students have chosen option (b), i.e., they prefer as roommates anyone compared to someone from the preference lists of their preferred friends.  The reasons they have provided for this decision, the number of students who have given the same reason for their decisions before the interview and changed their minds during/after the interview are also provided. }
\label{tab:reasonB}
\centering
{\tablefont\begin{tabular}{@{\extracolsep{\fill}}lcc}
\topline
 & \multicolumn{2}{c}{\#Students}  \\
\cline{2-3}
Reasons for choosing option (b) & before interview (6) & during/after interview (2) \midline
my friend may have friends I don't like &  2 & 0    \\ \cline{1-3}
my friend's friend is also my friend,   &    & \\
I could include them in the list if I wanted               &  1 & 0    \\ \cline{1-3}
at least match with someone             &  1 & 0 	\\ \cline{1-3}
new people, new experience              &  1 & 1    \\ \cline{1-3}
clear boundaries, less emotional pressure                 & 1 & 1 \\ \hline \hline
\end{tabular}}
\end{table}




\section{Discussions}

Finding a solution to every \srti instance cannot be guaranteed, and, to address this limitation, we have introduced a knowledge-based method that extends the preference lists of agents with ``suitable" candidates. In our earlier work, we have introduced a method for identifying such suitable candidates by considering the living styles and habitual preferences of students (e.g., smoking, studying). However, it might not always be sufficient to find such suitable candidates because of relevant but inaccessible  knowledge about the students, due to ethical and fairness concerns. For instance, it would not be appropriate to ask questions in a roommate search questionnaire related to students' political opinions, although people with similar ideas tend to get along better. 
As also observed in our surveys and interviews, the networks of preferred friends implicitly provide such valuable insights into potential compatibility between roommates, leading to personalized good-enough matchings for roommates.

Motivated by the promising empirical results of our emerging application, we plan to enrich our method with further types of relevant knowledge and deploy it at our university. 

As mentioned in the introduction, when an \srti instance does not have a stable matching, alternative approaches ``relax'' the notion of stability (\eg almost stability, sticky-stability, popularity). Our approach of finding ``good-enough'' matchings by extending preference lists is orthogonal to these approaches. Investigations of combining these two different approaches is a part of our ongoing work.

\paragraph{\bf Acknowledgments.} We would like to thank the participants of the poll survey and interviews for their participation and valuable feedback, Baturay Yilmaz for helpful discussions, and the anonymous reviewers for their valuable comments.


\bibliographystyle{tlplike}

\newpage
\appendix
\section*{Appendix A}
Motivated by real-world applications, and considering that stable roommates problems do not always have solutions, we continue our studies to compute ``good-enough" matchings. In addition to the agents' habits and habitual preferences, we consider their networks of preferred friends, and introduce a method to generate personalized solutions to stable roommates problems.

We investigate the usefulness of utilizing the knowledge about the networks of the agents' preferred friends from the perspective of the users. We design and perform a poll survey and interviews with students to obtain their feedback, and use subjective measures to evaluate the results. The results of the survey and interviews are presented in the main article.

In this supplementary material, we present mode detailed information about our interviews that are conducted to better understand the responses given in our poll survey.

Each interview started with the presentation of the following text to the participant, to inform them about the purpose of the interview and to get their permission for recording.

\smallskip
{\it Hi! Thank you for your time for the interview. The purpose of this interview is to understand the reasons for your answer you gave in our poll survey. If you do not remember the question, I can remind you or you can read it here. I will take notes during the interview, but to make sure I do not miss anything from your responses, I would like to record your interview. The recording will not be shared with anyone outside, and your identity will remain anonymous. Before we begin, do you give the permission for recording your interview?}
\smallskip

After that, the interviews proceeded as a dialogue, as illustrated in the following two examples (obtained from the video recordings of the interviews, with the permission of the participants).

\medskip
{\noindent \bf Interview with Volunteer~3 (male)} \\
Interviewer: You have already filled out the poll survey. What did you answer? \\
\underline{Volunteer~3:} I chose ``It does not matter, anyone will do.". \\
Interviewer: Why did you choose this one? \\
\underline{Volunteer~3:} My friend on my list has friends that I do not get along with.  \\
Interviewer: Do you have to know your friends' friends? \\
\underline{Volunteer~3:} Yes, I generally know them. I actually like them, but I do not want to be a roommate with them. For example, they are smoker, and I do not want a smoker roommate. Such a small problems become a huge problem within the social group. That's why there may be some friends of my friend I don't want to become roommate with. \\
Interviewer: What if the dormitory system mentioned that you can also specify students you do not want to be matched with as roommates, would you change your decision? \\
\underline{Volunteer~3:} Now, my choice will change.  \\
Interviewer: Why did you change your mind?\\
\underline{Volunteer~3:} If I can specify people who do not want to become a roommate, then assigning a friend of my friend is better than anyone random. Sincerity is important while staying together. \\
Interviewer: Can you say a little more about?\\
\underline{Volunteer~3:} There exist at least one common thing: our mutual friend. Common things make people closer to each other. \\
Interviewer: What kinds of things, can you say it more? \\
\underline{Volunteer~3:} Personality traits. Since similar personality traits of friends are similar, we may get along well.  \\
Interviewer: Consider again the same scenario in the poll survey, let's assume that the dormitory system also asks you for your preferences about smoking and sleeping habits. If this system cannot match you with any of your friends in your preference list, would you prefer to roommate with someone based on your preferences about these habits or anyone? \\
\underline{Volunteer~3:} I prefer a roommate who is closer to my preferences. \\
Interviewer: Under this setting, which one do you prefer: someone from preference lists of your friends or someone with whom you have similar preferences in terms of these criteria?\\
\underline{Volunteer~3:} Again, I prefer a roommate who is closer to my habitual preferences.\\
Interviewer: Why? \\
\underline{Volunteer~3:} Actually, too much connection leads to separation. That's why I prefer a roommate without connection in this case. In fact, the habitual similarity also shows the similarity of their characteristic traits. So, it is better than friends of my friend, since it is only shows the similarity of our characteristic traits. \\
Interviewer: Thank you. Is there anything you want to add?\\
\underline{Volunteer~3:} No, thanks. \\

{\noindent \bf Interview with Volunteer~6 (female)} \\
Interviewer: You have already filled out the poll survey. What did you answer? \\
\underline{Volunteer~6:} I chose ``Yes, I prefer this.". \\
Interviewer: Why did you choose this one? \\
\underline{Volunteer~6:} I think my preferred friends specify their preference lists similar with me. That's why I think I can get along well with their preferred friends. \\
Interviewer: Why did you think like that?\\
\underline{Volunteer~6:} In many ways, but first of all it is important to be clean. Also, being quiet or not turning on the light at certain hours. Apart from that, she tells us when she has an online meeting in our room, I will pay attention during her meeting. Likewise, she does the same for me. Respect is the key point I guess. Being able to have a good conservation is also very important. \\
Interviewer: So in what ways do you think you will have a good conservation? \\
\underline{Volunteer~6:} I do not have very high expectations. At least say ``Good morning" in the morning.  \\
Interviewer: I guess you've experienced something about this, right? \\
\underline{Volunteer~6:} Yes, with my first roommate. She did not talk at all. There were no communication in the room. It was very bad experience. Even when I said ``hello", I could not get a reply. \\
Interviewer: Did you match randomly with her? \\
\underline{Volunteer~6:} Yes. Actually, randomly assignment idea is good for meeting new people, but sometimes it can feel bad. \\
Interviewer: Do you have experience being roommates with your friend's friend? \\
\underline{Volunteer~6:} Yes, we matched randomly again, but we learned that we had a mutual friend. And we get along very well, we are happy. That's why there is a positive atmosphere in the room. We spend time together not only in our room but also outside, and participate in some events. \\
Interviewer: You have a roommate experience where you are randomly matched, and you also have a roommate experience where you are randomly matched but have a mutual friend. Can you compare them? \\
\underline{Volunteer~6:} I actually have 3 random roommate experience. As I mentioned before, the first one was awful. Then, I had a foreign roommate, and it was a lot of fun. I learned lots of new things from her and her culture. But, we got along even better with my current roommate, who is the one we have common friend. This is really good experience for me. We have a very good connection, and we can share our issues. We can spend time together. I think people have good communication when they have common friends. Thanks to this, we became much closer I guess.  \\
Interviewer: Consider again the same scenario in the poll survey, let's assume that the dormitory system also asks you for your preferences about smoking and sleeping habits. If this system cannot match you with any of your friends in your preference list, would you prefer to roommate with someone based on your preferences about these habits or anyone? \\
\underline{Volunteer~6:} Based on my preferences. \\
Interviewer: Under this setting, which one do you prefer: someone from preference lists of your friends or someone with whom you have similar preferences in terms of these criteria?\\
\underline{Volunteer~6:} In this situation, I prefer friends of my friends. Since there are other good things in addition to similar habitual preferences. On the other hand, I may become a roommate based on our habitual preferences, but she may be someone who is cold, and uncommunicative. So, friend of my friend is better option.  \\
Interviewer: Thank you. Is there anything you want to add?\\
\underline{Volunteer~6:} No, thanks. \\

\end{document}